\documentclass{article}
\usepackage{ijcai18}
\usepackage{graphicx}
\usepackage{times}
\usepackage{xcolor}
\usepackage{soul}
\usepackage[utf8]{inputenc}
\usepackage[small]{caption}
\usepackage{blindtext}
\usepackage{url}
\usepackage{amsfonts}
\usepackage{amsmath}
\usepackage{algorithm}
\usepackage{algorithmicx}
\usepackage{algpseudocode}

\title{Transfer Learning versus Multi-agent Learning \\
	regarding Distributed Decision-Making in Highway Traffic}

\author{
Mark Schutera$^{1, 4}$, 
Niklas Goby$^{2, 3}$, 
Dirk Neumann$^{2}$, 
Markus Reischl$^{1}$
\\ 
$^1$ Institute for Automation and Applied Informatics, Karlsruhe Institute of Technology \\
$^2$ Chair for Information Systems Research, University of Freiburg\\
$^3$ IT Innovation Chapter Data Science, ZF Friedrichshafen AG\\
$^4$ Research and Development, ZF Friedrichshafen AG\\
mark.schutera@kit.edu,
niklas.goby@is.uni-freiburg.de,
dirk.neumann@is.uni-freiburg.de,
markus.reischl@kit.edu
}

\begin{document}

\maketitle

\begin{abstract}
  Transportation and traffic are currently undergoing a rapid increase in terms of both scale and complexity. At the same time, an increasing share of traffic participants are being transformed into agents driven or supported by artificial intelligence resulting in mixed-intelligence traffic. This work explores the implications of distributed decision-making in mixed-intelligence traffic. The investigations are carried out on the basis of an online-simulated highway scenario, namely the MIT \emph{DeepTraffic} simulation. In the first step traffic agents are trained by means of a deep reinforcement learning approach, being deployed inside an elitist evolutionary algorithm for hyperparameter search. The resulting architectures and training parameters are then utilized in order to either train a single autonomous traffic agent and transfer the learned weights onto a multi-agent scenario or else to conduct multi-agent learning directly. Both learning strategies are evaluated on different ratios of mixed-intelligence traffic. The strategies are assessed according to the average speed of all agents driven by artificial intelligence. Traffic patterns that provoke a reduction in traffic flow are analyzed with respect to the different strategies.
\end{abstract}

\section{Introduction}
The level of automation in traffic and transportation is increasing rapidly,  especially in the context of highway scenarios, where complexity is reduced in comparison to urban street scenarios. Traffic congestion is annoying, stressful, and time-consuming. Progress in the area of autonomous driving thus offers the opportunity to: Improve this condition, enhance traffic flow, and yield corresponding benefits such as reduced energy consumption~\cite{WINNER.2015}. At the same time, autonomous systems are distributed by a number of different manufacturers and suppliers. This leads to the challenge of the interaction between different autonomous systems and human-operated vehicles. Therefore, it seems to be within the realm of possibility that increased automation in traffic may compromise the average flow of mixed intelligence traffic. As highway traffic can be described in terms of a multi-agent system with independent agents cooperating and competing to achieve an objective the key to high-performance highway traffic flow might lie within multi-agent learning and thus within the understanding and exploration of distributed decision-making and its strategies. Transfer learning is used with increasing frequency within deep learning and might prove able to adapt artificial neural networks to bordering tasks~\cite{IAI.2018}. Within the automotive industry, the pros and cons of each such strategy are still subject to ongoing discussions. This work contributes to this discussion by investigating the performance of transfer learning, as opposed to multi-agent learning, regarding distributed decision-making in highway traffic. For the experiments, agents are trained with different learning strategies and deploy them to the \emph{DeepTraffic} micro-traffic simulation, which was introduced along with the MIT 6.S094: Deep Learning for Self-Driving Cars course~\cite{FRIDMAN.2018}.  The aim of this study is to examine the impact on mixed intelligence traffic in the form it's expected to take with the adoption of Level 5 autonomous driving. To this end the subsequent steps are taken:
\begin{itemize}
	\item Traffic agents are trained within a micro-traffic simulation, through deep reinforcement learning. 
	\item An evolutionary algorithm is designed to embed the traffic agents' learning procedure.
	\item A single traffic agent's model is applied to multiple agents (transfer learning strategy).
	\item Multiple traffic agents are jointly trained (multi-agent learning strategy).
	\item The two learning strategies are evaluated by means of speed and traffic flow patterns.
\end{itemize}

\section{Micro-Traffic Simulation Environment}

In the DeepTraffic\footnote{\url{https://selfdrivingcars.mit.edu/deeptraffic/}} challenge, the task is to train a car agent with the goal of achieving the highest average speed over a period of time. In order to succeed, the agent has to choose the optimal action $a_t$ at each step in time given the state $s_t$. Possible actions are: \emph{accelerate}, \emph{decelerate}, \emph{goLeft} (lane change to the left), \emph{goRight} (lane change to the right) or \emph{noAction} (remain in the current lane at the same speed). The agent's observed state $x_t$ at time step $t$ is defined as the number of grid cells surrounding the agent. The size of the slice is adjustable via three different parameters: \emph{lanesSide}, representing the width of the slice; \emph{patchesAhead}, denoting the length of the slice in the forward direction; and \emph{patchesBehind}, representing the length of the slice in the backward direction. Depending on the parameter \emph{temporal\_window}, $w$, the state $s_t$ can be transformed into a sequence $s_t = x_{t-w}, a_{t-w}, x_{t-w+1}, a_{t-w+1}, \dots, x_{t-1}, a_{t-1}, x_{t}, a_{t}$. If $w=0$, then $s_t=x_t$. Cell values denote the maximum speed the agent can achieve when it is inside the cell. The maximum speed in an empty cell is set to $80$ mph. A cell occupied by a car maintains the speed of the car. 

There are a total of $20$ cars inside the environment, for which the intelligent control of up to $11$ cars is allowed. The remaining cars choose their actions randomly.
Central to the intelligent control is a neural network. It receives the observed state $s_t$ as input and returns an action $a_t$, therefore functioning as the agent's behavior. The implemented algorithm is a JavaScript implementation of the famous DQN \cite{Mnih.2013} algorithm. Please refer to section~\ref{drl-dqn} for more information.

The environment allows for the adjustment of a whole set of hyperparameters in order to push the agents' performance. Table~\ref{tab:Hyperparameter} lists the most important hyperparameters, which have proven to have a significant influence on the agents' performance \cite{FRIDMAN.2018}.
These hyperparameters, as well as the network architecture itself, can be directly adjusted within the browser. To automate the configuration, training, and validation process for the experiments a Python-based helper robot using the Selenium\footnote{\url{http://selenium-python.readthedocs.io/}} package was implemented.

\section{Training Advanced Traffic Agents}

\subsection{Deep Reinforcement Learning and the Deep Q-Network (DQN)}
\label{drl-dqn}

Deep reinforcement learning (DRL) is the combination of two general-purpose frameworks: reinforcement learning (RL) for decision-making, and deep learning (DL) for representation learning \cite{Silver.2016}.

In the RL framework, an agent's task is to learn actions within an initially unknown environment. The learning follows a trial-and-error strategy based on rewards or punishments. The agent's goal is to select actions that maximize the cumulative future reward over a period of time. 
In the DL framework, an algorithm learns a representation from raw input that is required to achieve a given objective.  The combined DRL approach enables agents to engage in more human like learning whereby they construct and acquire their knowledge directly from raw inputs, such as vision, without any hand-engineered features or domain heuristics. 
This new generation of algorithms has recently achieved human like results in mastering complex tasks with a very large state space and with no prior knowledge \cite{Mnih.2013,Mnih.2015,Silver.2017}. 

The simulation environment, per default, implements a DQN algorithm introduced in \cite{Mnih.2013,Mnih.2015} for training the advanced traffic agents. As a variant of the popular Q-learning \cite{Watkins.1992} algorithm, DQN uses a neural network to approximate the optimal state-action value function (i.e. $Q$-function). To make this work, DQN utilizes four core concepts: experience replay \cite{Lin.1993}, a fixed target network, reward clipping, and frame skipping \cite{Mnih.2015}. 

The resulting approximate state-action value function $Q(s,a,\theta_i)$ is parametrized through $\theta_i$, in which $\theta_i$ are the parameters (i.e weights) of the Q-network at iteration $i$ \cite{Mnih.2015}. To train the Q-network at iteration $i$, one has to minimize the following loss function:

\begin{equation}
L_i(\theta_i) = \mathbf{E}_{(s,a,r,s^{\prime}) \sim U(\mathcal{D})} \left[\left(y_i - Q(s,a; \theta_i)\right)^2  \right], 
\end{equation}
in which $(s,a,r,s^{\prime}) \sim U(\mathcal{D})$ represents samples of experiences, drawn uniformly at random from the experience replay memory $\mathcal{D}$  (experience replay), $y_i = r + \gamma \max_{a^{\prime}} Q(s^{\prime},a^{\prime}; \theta_{i}^{-})$ is the target for iteration $i$, $\gamma$ is the discount factor determining the agent's horizon, $\theta_{i}$ are the parameters of the Q-network at iteration $i$ and $\theta_{i}^{-}$ are the network parameters used to compute the target at iteration i, which updates every $C$ steps and hold fix otherwise (fixed target network) \cite{Mnih.2015}. Algorithm~\ref{alg} outlines the full pseudo-code algorithm.

\begin{algorithm}[t]
	\begin{algorithmic}
		\State Initialize replay memory $\mathcal{D}$ to capacity $N$
		\State Initialize action-value function $Q$ with random weights
		\For{episode $=1,M$} 
		\State Observe initial state $s_1$ 
		\For {$t=1,T$}
		\State With probability $\epsilon$ select a random action $a_t$
		\State otherwise select $a_t = \max_{a} Q^*(s_t, a; \theta)$
		\State Execute action $a_t$ 
		\State Observe reward $r_t$ and state $s_{t+1}$
		\State Store experience $\left(s_t, a_t, r_t, s_{t+1}\right)$ in $\mathcal{D}$
		\State Sample random minibatch of transitions $\left(s_j,a_j,r_j,s_{j+1}\right)$ from $\mathcal{D}$
		\State Set
		$y_j =
		\left\{
		\begin{array}{l l l l}
		\text{for terminal } s_{j+1}: \\
		r_j  \\
		\text{for non-terminal } s_{j+1}: \\
		r_j + \gamma \max_{a'} Q(s_{j+1}, a'; \theta)
		\end{array} \right.$
		\State Train the Q network using $\left(y_j - Q(s_j, a_j; \theta) \right)^2$ as loss
		\EndFor
		\EndFor
	\end{algorithmic}
	\caption{Deep Q-learning with Experience Replay}
	\label{alg}
\end{algorithm}

\subsection{Extended Hyperparameter Search}
\label{ch:Hyperparametersearch}
Within deep reinforcement learning there arises the need for a structured approach to determine suitable hyperparameter configurations $\lambda$. This is important for both the neural network's architecture and the training process. The following approach fulfills this requirement over multiple search iterations. The micro-traffic simulation has already been used to conduct a large-scale, crowd-sourced hyperparameter search~\cite{FRIDMAN.2018}. In a first step, the proposals drawn from this hyperparameter search are utilized in order to define the intervals of the hyperparameters (see Tab.~\ref{tab:Hyperparameter}). Building on the hyperparameter bounds, a $15$-fold random search is performed, as proposed by~\cite{RANDOM.2012,GOODFELLOW.2016}.  

\begin{table}[ht]
	\centering
	\caption{Itemization of the hyperparameter search space. Lower and upper bounds of the hyperparameter configuration $\lambda$. Configuration depicts the optimal hyperparameter configuration found with the presented hyperparameter search.}
	\begin{tabular*}{\columnwidth}{p{.4\columnwidth}p{.2\columnwidth}p{.4\columnwidth}}
		\textbf{Hyperparameter}     & \textbf{Interval} & \textbf{Configuration} \\
		lanesSide					& $[3; 6]$      &					3\\
		patchesAhead 				& $[1; 55]$     &					30\\
		patchesBehind 				& $[1; 20]$     &					13\\
		trainIterations ($M$) 		& $[10k, 100k]$ &					100k\\
	    temporal window 			& $[0]$         &					0\\
		num neurons					& $[1; 100]$    &					21\\
		learning rate ($\alpha$)    & $[0.0001; 0.1]$ &					0.00017\\
		momentum					& $[0;1]$ )     &					0.57\\
		batch size 					& $[1; 128]$    &					53\\
		l2 decay 					& $[0.01]$      &					0.01\\
		experience size ($N$) 		& $[3k; 10k]$   &					5000\\
		start learn threshold 		& $[500]$       &					500\\
		gamma ($\gamma$)			& $[0.8; 1]$    &					0.9\\
		learning steps total 		& $[10k; ti]$ &					54129\\
		learning steps burnin 		& $[1k; ti/2]$&					1083\\
		epsilon min 				& $[0; 1]$ &					0.86\\
		epsilon test time 			& $[0; 1]$ &					0.22\\
		number of layers 			& $[4;7]$  &					7\\
	\end{tabular*}
	\label{tab:Hyperparameter}
\end{table}

Subsequently, the five best performing networks, which are generated by the random search were utilized to initialize an elitist evolutionary algorithm. The hyperparameter search for artificial neural networks is inhibited by the comparatively long training time for each hyperparameter configuration. Therefore, an elitist fast converging evolutionary algorithm was deployed to automate the process further. The whole hyperparameter search process reduces the effects of bad agent configuration, rendering the effects of transfer learning and multi-agent approaches more visible and reproducible. In the future, we would also like to exploit the hyperparameter tuning capabilities of evolutionary algorithms to create highly optimized agents \cite{Salimans.2017,Such.2017,Conti.2017}.

\section{Learning Strategies for Systems Based on Distributed Decision-Making}

\subsection{Transfer Learning}
\begin{figure}[ht]
	\centering
	\includegraphics[width=2.9in]{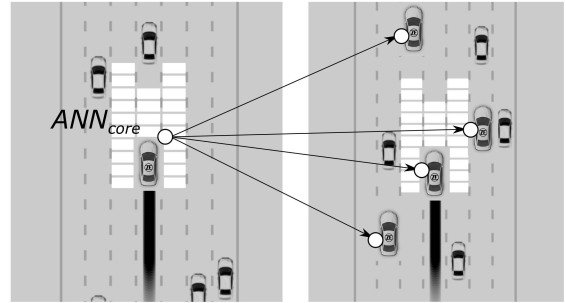}
	\caption{Two screenshots from the micro-traffic simulation. The highlighted cells depict the catchment areas of the safety system, which automatically slows down the car to prevent collisions. The vehicles with the logo represent trainable agents, while those without a logo are not trainable and exhibit random behavior. The left figure shows the training process of a core network ANN$_{core}$, whereas on the right figure illustrates the pretrained core network being deployed among multiple agents.}
	\label{fig:TransferLearning}
\end{figure}

Throughout the transfer learning strategy, a first core neural network ANN$_{core}$ is trained. The network is trained with a single agent deployed within the micro-traffic simulation. The training is iterated while training and evaluating different hyperparameter configurations (for hyperparameter configuration see Tab.~\ref{tab:Hyperparameter}).     
Subsequently, the learned model is repurposed for a multi-agent system.
The decision-making process is distributed over independent, multiple agents. The transfer learning approach presented here, is based on parameter sharing among multiple agents while the agents maintain their ability to carry out self-determined actions. To that end, the previously learned weights of ANN$_{core}$ are transfered onto a second, third, and so on agent $A_i$, as described by~\cite{TRANSFER.2009} and displayed in Fig.~\ref{fig:TransferLearning}. 

\subsection{Multi-agent Learning}
\begin{figure}[ht]
	\centering
	\includegraphics[width=2.7in]{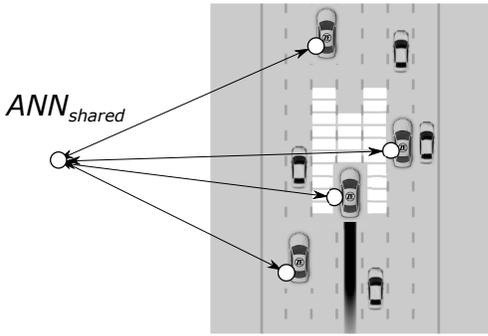}
	\caption{The network ANN$_{shared}$ is shared across multiple agents $A_i$ and trained with respect to a reward function that depends on the outcome of all agents' actions and their influence on the average speed $\bar{v}$.}
	\label{fig:MultiAgentLearning}
\end{figure}

Within the multi-agent learning strategy, the agents are trained simultaneously without being aware of each other. More precisely, they have to interact with each other without the possibility to communicate among themselves.  This makes joint planning impossible. The resulting network ANN$_{shared}$ is trained with the joint objective of achieving the average speed for all agents, but as in the transfer learning scenario, actions are taken individually and in a greedy way. The neural network's ANN$_{shared}$ parameters are distributed and shared across all agents $A_i$ (see Fig.~\ref{fig:MultiAgentLearning}).   In contrast to the transfer learning approach, the multi-agent strategy enables the agents to learn to interact directly with other agents in order to increase the reward~\cite{MULTIAGENT.2012}.   

\subsection{Traffic Pattern Annotation}
\label{ch:trafficpatternannotation}
\begin{figure}[ht]
	\centering
	\includegraphics[width=2.9in]{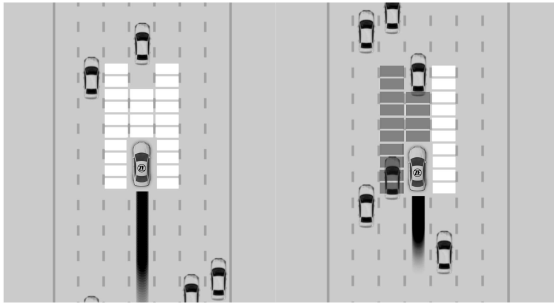}
	\caption{Two screenshots from the micro-traffic simulation. The highlighted cells depict the catchment areas of the safety system, which automatically slows down the car to prevent collisions. The left figure shows the vehicle in a state of free passage, whereas in the right figure, the left lane and area directly in front of the vehicle are blocked.}
	\label{fig:CongestionPatternVector}
\end{figure}

In order to summarize and to make traffic flow analyzable an annotation for traffic patterns is introduced. As traffic flow can be defined as the absence of traffic congestions, the proposed traffic pattern annotation is based on analyzing congestion patterns (see Fig.~\ref{fig:CongestionPatternVector}).

The congestion pattern (see Fig.~\ref{fig:CongestionPatternVector}) is cast into a feature vector annotation $\textbf{cp}$
\[
\textbf{cp} = \{ B, S, D, C \}.
\] 
$\textbf{cp}$ comprises a boolean $B$ stating whether the car is blocked to the front and/or sides ($1$) or whether one of the lanes -- left lane, front lane, or right lane -- is passable ($0$) due to the safety regulations within the safety catchment area. Furthermore, the feature vector annotation takes into account  the speed $S$ at which the agent drove into the congestion as well as the loss in speed or deceleration $D$ the agent's vehicle experiences within half a second in simulated time after encountering the congestion. The feature $C$ reflects whether the agent was compromised by another intelligent agent and thus assesses the amount of cooperation during evaluation. The number of congestion throughout the evaluation runs is taken into account as $n_{cp}$.

\section{Experiments}

The first experiments focus on a hyperparameter search as described in section~\ref{ch:Hyperparametersearch}. The hyperparameter configuration for the elitist evolutionary algorithm are as follows:
A small population size, $\mu = 5$, and a directed population initialization  by means of a random search keeping the best parent during transition into the next generation. The crossover rate is set to $p_{cross} = 0.3$ and the mutation rate to $p_{mut} = 0.1$. This approach significantly favors exploitation over the exploration of the hyperparameter space. Hence, the approach converges in short time while exhibiting the disadvantage of reduced exploration of the hyperparameter space.        

In order to compare the transfer learning strategy to the multi-agent strategy (see Fig.~\ref{fig:PerformanceDevelopment}), the neural network architecture and training parameters (see Tab.~\ref{tab:Hyperparameter}) discovered by the hyperparameter configuration search are further utilized. Each strategy is applied to different numbers of trainable agents, ranging from $1$ up to $11$ agents. Each arrangement is evaluated $5$-fold to meet expected deviations due to differing evaluation data. However, this is found to pose only a minor issue as the minimal and maximal validation performance for each arrangement spans less than $0.5$ mph in all arrangements.

\subsection{Results}

\begin{figure}[ht]
	\centering
	\includegraphics[width=3in]{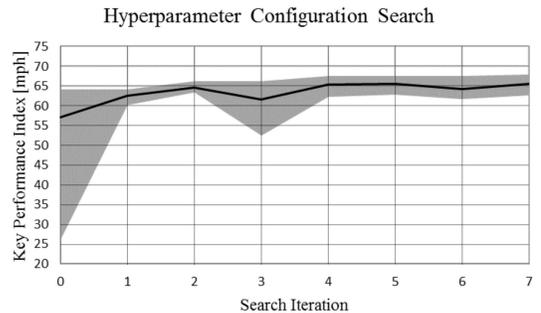}
	\caption{Development of the configurations' performance in [mph] over the search iterations. Iteration $0$ depicts the random search approach, while iteration $1$ shows the sampling of the top five configurations from the random search. Iterations $2$-$7$ represent the generations of the evolutionary algorithm. The upper bound of the graph is the maximum performance, the lower bound is the minimum performance, and the black line represents the mean performance in the respective search iteration.}
	\label{fig:HyperparameterSearch}
\end{figure}

In the quest to find a high-performance hyperparameter configuration, the $15$-fold random search makes a start by evaluating in the  micro-traffic simulation configurations reaching a maximum average speed of $64.13$~mph (see search iteration $0$ in Fig.~\ref{fig:HyperparameterSearch}). The average speed is used as an indicator for traffic flow. The best five configurations are selected to initialize the evolutionary algorithm which leaves the configurations with the maximum at $64.13$~mph, the minimum at $60.27$~mph, and the mean at $62.55$~mph. The evolutionary algorithm is deployed over six generations (see search iterations $2$-$7$ in Fig.~\ref{fig:HyperparameterSearch}). As discussed in Section \ref{ch:Hyperparametersearch}, the evolutionary algorithm is elitist with a focus on exploitation and enabling an extended exploration. However, the influence of exploration is observed in search iteration $3$, while the stronger exploitation is evident in search iteration $4$, where the range of values is again decreased. After completion of the evolutionary algorithm, the configurations have a maximum of $67.93$~mph, a minimum of $62.63$~mph and a mean of $65.66$~mph. This shows an increase of $3.8$~mph without any user interaction apart from choosing educated upper and lower bounds for the hyperparameter search space.                    

\begin{figure}[ht]
	\centering
	\includegraphics[width=3in]{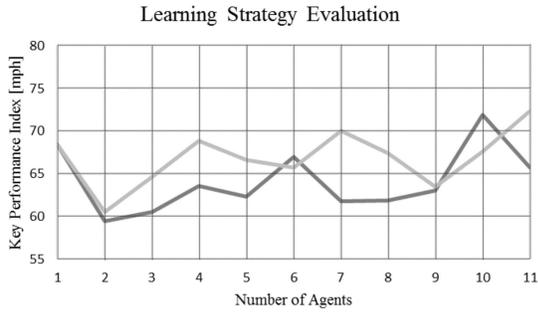}
	
	\caption{Evaluation of the key performance index [mph] for the various numbers of agents deployed in the micro-traffic simulation. The transfer learning strategy is shown in dark gray, while the multi-agent learning strategy is shown in light gray.}
	
	\label{fig:PerformanceDevelopment}
\end{figure}

Both strategies experience a drop in performance when applied to multi-agent scenarios (see Fig.~\ref{fig:PerformanceDevelopment}). As for the initial addition of supplementary agents the performance downturn is likely due to the fact that the network architecture and training hyperparameters have been optimized according to a desirable single agent performance which then faces a different scenario during the reconditioned evaluation.          
Notwithstanding an overall increase in performance, associated with an increase in the number of agents can be recognized. The slopes of the regression curves are: $0.342$ mph per agent added for the transfer learning strategy and  $0.379$ mph per agent added for the multi-agent training strategy (compare with Fig.~\ref{fig:PerformanceDevelopment}). The multi-agent strategy, having the edge over the transfer learning strategy, is able to profit from training in a multi-agent scenario. By contrast, agents in the transfer learning strategy never had the opportunity to learn how to react to and interact with other trained agents. 

\begin{figure}[ht]
	\centering
	\includegraphics[width=3in]{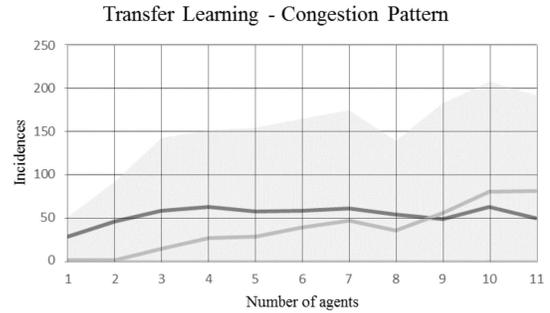}
	\centering
	\includegraphics[width=3in]{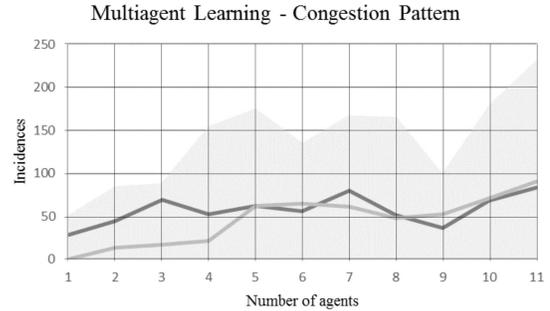}
	
	\caption{Representation of the amount of congestions (incidences) for the various numbers of agents deployed in the micro-traffic simulation. On top, the transfer learning strategy and at the bottom the multi-agent learning strategy is shown. The gray area is the absolute amount of congestions detected during evaluation. The light gray line depicts the amount of congestions with participation of another trained agent. The dark gray line depicts the amount of congestions in which the ego-vehicle is fully blocked on all lanes.}
	
	\label{fig:CongestionDevelopment}
\end{figure}
 
Further insight is gained by analyzing the traffic congestion feature vectors (see Section~\ref{ch:trafficpatternannotation} and Fig.~\ref{fig:CongestionDevelopment}). What strikes one the most is the counterintuitive finding that the number of congestions (gray area) increases with the amount of trained agents deployed in the micro-traffic simulation and as the average evaluation speed increases. Simultaneously, the number of congestions in which the car is held in full enclosure (dark gray line) remains constant, fluctuating around $50$ incidences. 

\begin{figure}[ht]
	\centering
	\includegraphics[width=3in]{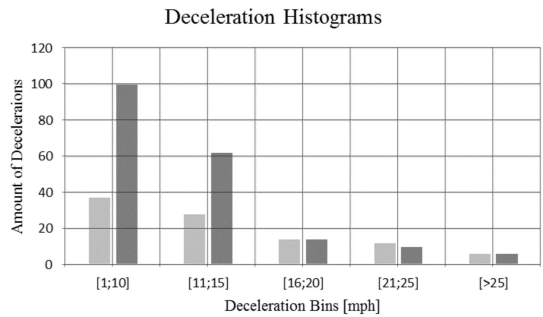}
	
	\caption{Representation of the number of congestions divided into $5$ mph bins, which outline the magnitude of deceleration. Light gray bars are associated with the transfer learning with a single agent, whereas the dark gray bars are associated with the transfer learning with eleven agents.}
	
	\label{fig:Histogram}
\end{figure}

This is only seemingly contradictory, as the largest part of the increased number of congestion incidences may be attributed to low decelerations $1$-$11$ (see Fig.~\ref{fig:Histogram}). In terms of traffic flow, this means that the trained agents are able to anticipate and withdraw from potentially congestive positions in advance or else dissolve a formation conducive to congestion. Thus, the trained agents are able to accelerate again shortly after driving into an area of congestion which leads to better performance.

\section{Conclusion}

The influence of transfer learning and multi-agent learning in the presence of multiple trainable agents, has been investigated with respect to distributed decision-making in order to increase simulated highway traffic flow.
Both strategies were implemented and evaluated in the micro-traffic simulation environment. Since the micro-traffic simulation only allows for multi-agent learning the newly conducted strategy-comparison and deployment of the transfer learning strategy as well as the evaluation tooling allows for extended testing and evaluation.   

It was demonstrated, that transfer learning strategies are applicable within the utilized micro-traffic simulation. A beneficial effect of such strategies correlating with the amount of trainable agents deployed in mixed-intelligence traffic has been shown. It was found that the transfer learning strategy and the multi-agent strategy were reaching approximately the same level of performance, while also displaying similar characteristics. Concentrating on traffic patterns, it became evident that the number of congestions an agent experiences not necessarily contingent on the average speed. More important are the magnitude of deceleration required of the agent and the time needed to withdraw from a congested situation. The micro-traffic scenario is a vast simplification of real traffic. Our findings suggest that multi-agent learning has an edge with respect to performance in scenarios with more intelligent agents involved. This leads to the assumption that with a growing number of intelligent agents taking to the roads, multi-agent learning strategies will be inevitable.       

Further comparisons between the investigated multi-agent strategies might reveal explicit distinctions. To this end, investigating ratios with a higher share of trainable agents is advisable. Moreover, the multi-agent strategies should benefit from a network architecture and training design that is tailored with respect to the multi-agent scenario (as opposed to the single-agent scenario). Increasing the amount of training iterations and deepening the hyperparameter search is recommended.            

\section*{Acknowledgments}

We thank Dr. Jochen Abhau and Dr. Stefan Elser from Research and Development,  as well as the whole Data Science Team at ZF Friedrichshafen AG, for supporting this research. Thank you for all the assistance and comments that greatly improved this work. We would also like to express our gratitude to Prof. Dr. Ralf Mikut from the Institute for Automation and Applied Informatics, Karlsruhe Institute of Technology, who provided insight and expertise that greatly enhanced this and other research.

\bibliographystyle{named}
\bibliography{deep_traffic}

\begin{thebibliography}{}

\bibitem[\protect\citeauthoryear{Bergstra and Bengio}{2012}]{RANDOM.2012}
James Bergstra and Yoshua Bengio.
\newblock Random search for hyper-parameter optimization.
\newblock {\em Journal of Machine Learning Research}, 13(Feb):281--305, 2012.

\bibitem[\protect\citeauthoryear{Conti \bgroup \em et al.\egroup
  }{2017}]{Conti.2017}
Edoardo Conti, Vashisht Madhavan, Felipe~Petroski Such, Joel Lehman, Kenneth~O.
  Stanley, and Jeff Clune.
\newblock Improving exploration in evolution strategies for deep reinforcement
  learning via a population of novelty-seeking agents.
\newblock {\em CoRR}, abs/1712.06560, 2017.

\bibitem[\protect\citeauthoryear{Fridman \bgroup \em et al.\egroup
  }{2018}]{FRIDMAN.2018}
Lex Fridman, Benedikt Jenik, and Jack Terwilliger.
\newblock Deeptraffic: Driving fast through dense traffic with deep
  reinforcement learning.
\newblock {\em CoRR}, abs/1801.02805, 2018.

\bibitem[\protect\citeauthoryear{Goodfellow \bgroup \em et al.\egroup
  }{2016}]{GOODFELLOW.2016}
Ian Goodfellow, Yoshua Bengio, and Aaron Courville.
\newblock {\em Deep Learning Book}.
\newblock MIT Press, 2016.
\newblock \url{http://www.deeplearningbook.org}.

\bibitem[\protect\citeauthoryear{Lin}{1993}]{Lin.1993}
Long-Ji Lin.
\newblock Reinforcement learning for robots using neural networks.
\newblock Technical report, Carnegie-Mellon Univ Pittsburgh PA School of
  Computer Science, 1993.

\bibitem[\protect\citeauthoryear{Mnih \bgroup \em et al.\egroup
  }{2013}]{Mnih.2013}
Volodymyr Mnih, Koray Kavukcuoglu, David Silver, Alex Graves, Ioannis
  Antonoglou, Daan Wierstra, and Martin Riedmiller.
\newblock Playing atari with deep reinforcement learning.
\newblock {\em arXiv preprint arXiv:1312.5602}, 2013.

\bibitem[\protect\citeauthoryear{Mnih \bgroup \em et al.\egroup
  }{2015}]{Mnih.2015}
Volodymyr Mnih, Koray Kavukcuoglu, David Silver, Andrei~A Rusu, Joel Veness,
  Marc~G Bellemare, Alex Graves, Martin Riedmiller, Andreas~K Fidjeland, Georg
  Ostrovski, et~al.
\newblock Human-level control through deep reinforcement learning.
\newblock {\em Nature}, 518(7540):529, 2015.

\bibitem[\protect\citeauthoryear{Olivas \bgroup \em et al.\egroup
  }{2009}]{TRANSFER.2009}
Emilio~Soria Olivas, Jose David~Martin Guerrero, Marcelino~Martinez Sober, Jose
  Rafael~Magdalena Benedito, and Antonio Jose~Serrano Lopez.
\newblock {\em Handbook Of Research On Machine Learning Applications and
  Trends: Algorithms, Methods and Techniques - 2 Volumes}.
\newblock Information Science Reference - Imprint of: IGI Publishing, Hershey,
  PA, 2009.

\bibitem[\protect\citeauthoryear{Prodanova \bgroup \em et al.\egroup
  }{2018}]{IAI.2018}
N.~Prodanova, J.~Stegmaier, S.~Allgeier, S.~Bohn, O.~Stachs, B.~K{\"o}hler,
  R.~Mikut, and A.~Bartschat.
\newblock Transfer learning with human corneal tissues: An analysis of optimal
  cut-off layer.
\newblock {\em MIDL Amsterdam}, 2018.
\newblock Submitted paper, online available.

\bibitem[\protect\citeauthoryear{Salimans \bgroup \em et al.\egroup
  }{2017}]{Salimans.2017}
Tim Salimans, Jonathan Ho, Xi~Chen, and Ilya Sutskever.
\newblock Evolution strategies as a scalable alternative to reinforcement
  learning.
\newblock {\em arXiv preprint arXiv:1703.03864}, 2017.

\bibitem[\protect\citeauthoryear{Silver \bgroup \em et al.\egroup
  }{2017}]{Silver.2017}
David Silver, Thomas Hubert, Julian Schrittwieser, Ioannis Antonoglou, Matthew
  Lai, Arthur Guez, Marc Lanctot, Laurent Sifre, Dharshan Kumaran, Thore
  Graepel, et~al.
\newblock Mastering chess and shogi by self-play with a general reinforcement
  learning algorithm.
\newblock {\em arXiv preprint arXiv:1712.01815}, 2017.

\bibitem[\protect\citeauthoryear{Silver}{2016}]{Silver.2016}
David Silver.
\newblock {ICML 2016 Tutorial: Deep Reinforcement Learning}, 2016.

\bibitem[\protect\citeauthoryear{Such \bgroup \em et al.\egroup
  }{2017}]{Such.2017}
Felipe~Petroski Such, Vashisht Madhavan, Edoardo Conti, Joel Lehman, Kenneth~O.
  Stanley, and Jeff Clune.
\newblock Deep neuroevolution: Genetic algorithms are a competitive alternative
  for training deep neural networks for reinforcement learning.
\newblock {\em CoRR}, abs/1712.06567, 2017.

\bibitem[\protect\citeauthoryear{Tuyls and Weiss}{2012}]{MULTIAGENT.2012}
Karl Tuyls and Gerhard Weiss.
\newblock Multiagent learning: Basics, challenges, and prospects.
\newblock {\em Association for the Advancement of Artificial Intelligence},
  2012.

\bibitem[\protect\citeauthoryear{Watkins and Dayan}{1992}]{Watkins.1992}
Christopher~JCH Watkins and Peter Dayan.
\newblock Q-learning.
\newblock {\em Machine learning}, 8(3-4):279--292, 1992.

\bibitem[\protect\citeauthoryear{Winner \bgroup \em et al.\egroup
  }{2015}]{WINNER.2015}
Hermann Winner, Felix Lotz, Stephan Hakuli, and Christina Singer.
\newblock {\em Handbuch Fahrerassistenzsysteme - Grundlagen, Komponenten und
  Systeme f{\"u}r aktive Sicherheit und Komfort}.
\newblock Springer Vieweg, 3 edition, 2015.

\end{thebibliography}

\end{document}